\documentclass[10pt,twocolumn,letterpaper]{article}

\usepackage{iccv}
\usepackage{times}
\usepackage{epsfig}
\usepackage{graphicx}
\usepackage{amsmath}
\usepackage{amssymb}

\usepackage{url}

\usepackage{color}

\usepackage[utf8]{inputenc} %
\usepackage[T1]{fontenc}    %
\usepackage{booktabs}       %
\usepackage{amsfonts}       %
\usepackage{nicefrac}       %
\usepackage{microtype}      %
\usepackage{multirow}
\usepackage{algorithm}
\usepackage{algorithmic}
\usepackage{subcaption}
\usepackage{tablefootnote}
\renewcommand{\algorithmicrequire}{\textbf{Initialization:}}

\usepackage[pagebackref=true,breaklinks=true,letterpaper=true,colorlinks,bookmarks=false]{hyperref}

\iccvfinalcopy %

\def\iccvPaperID{0000} %

\begin{document}

\title{Network Automatic Pruning: Start NAP and Take a Nap}

\author{Wenyuan Zeng\\
University of Toronto\\
Uber ATG\\
{\tt\small wenyuan@cs.toronto.edu}
\and
Yuwen Xiong\\
University of Toronto\\
Uber ATG\\
{\tt\small yuwen@cs.toronto.edu}
\and
Raquel Urtasun\\
University of Toronto\\
Uber ATG\\
{\tt\small urtasun@cs.toronto.edu}
}

\maketitle

\begin{abstract}

Network pruning can significantly reduce the computation and memory footprint of large neural networks. To achieve a good trade-off between model size and performance, popular pruning techniques usually rely on hand-crafted heuristics and
require manually setting the compression ratio for each layer. This process is typically time-consuming and requires expert knowledge to achieve good results. In this paper, we propose NAP, an unified and automatic pruning framework for both {\it fine-grained} and {\it structured} pruning. It can find out unimportant components of a network and automatically decide appropriate compression ratios for different layers, based on a theoretically sound criterion. Towards this goal, NAP uses an efficient approximation of the Hessian for evaluating the importances of components, based on a Kronecker-factored Approximate Curvature method.
Despite its simpleness to use, NAP outperforms previous pruning methods by large margins. For fine-grained pruning, NAP can compress AlexNet and VGG16 by \textbf{25x}, and ResNet-50 by \textbf{6.7x} without loss in accuracy on ImageNet. For structured pruning (e.g. channel pruning), it can reduce flops of VGG16 by \textbf{5.4x} and ResNet-50 by \textbf{2.3x} with only 1\% accuracy drop.
More importantly, this method is almost free from hyper-parameter tuning and requires no expert knowledge. You can start NAP and then take a nap!

\vspace{-0.3cm}

\end{abstract}

\section{Introduction}
\vspace{-0.1cm}
Deep neural networks have proven to be very successful in many artificial intelligence tasks such as computer vision \cite{he2016deep, krizhevsky2012imagenet, simonyan2014very}, natural language processing \cite{sutskever2014sequence}, and robotic control \cite{mnih2013playing}. In exchange of such success, modern architectures are usually composed of many stacked layers parameterized with a large number of learnable weights.
As a consequence, modern architectures require considerable memory storage and intensive computation. For example, the VGG16 network \cite{simonyan2014very} has 138 million parameters in total and requires 16 billion floating point operations (FLOPs) to finish one forward-pass. This is problematic in applications that need to run on small embedded systems or that require low-latency to make decisions.

Fortunately, we can compress these large networks with little to no loss in performance by exploiting the fact that many redundancies exist within their parameters.
Several directions have been explored in the compression community such as quantization \cite{courbariaux2016binarized,gong2014compressing,rastegari2016xnor,wu2016quantized,zhu2016trained} and low-rank approximation \cite{gong2014compressing, jaderberg2014speeding, lebedev2014speeding}.
Among these directions, {\it pruning} techniques \cite{han2015deep,han2015learning, hassibi1993second, he2017channel, lecun1990optimal, wen2016learning} have been widely exploited due to their simplicity and efficacy. They aim at removing redundant and unimportant parameters from a network, and thus save storage and computation during inference.
More specifically, pruning techniques can be categorized into two branches. {\it Fine-grained pruning} \cite{han2015learning} removes individual parameters from a network, and thus saves the storage space and benefits for embedded systems.
{\it Structured pruning} \cite{he2017channel, luo2017thinet, wen2016learning}, on the other hand, typically removes whole channels (filters) from a network and thus achieves inference speed-up with no need of any hardware specialization.
In this paper, we follow the line of pruning techniques and propose an unified pruning approach that can be used to prune either individual parameters or channels.

The main challenges in pruning task are: 1) how to determine unimportant parameters (channels) in a layer, 2) how to decide the compression ratio for each layer in a network. Different pruning methods mainly differ on how they tackle these two problems. In the {\it fine-grained} domain, popular pruning methods typically rely on heuristics, such as weight magnitude to determine the importance of a parameter \cite{han2015deep, han2015learning}. They manually tune the compression ratios for different layers, and then prune the smallest number of parameters from each layer according to the compression ratios. However, a small magnitude does not necessarily mean an unimportant parameter \cite{hassibi1993second, lecun1990optimal}. Furthermore, because different layers (and different network architectures) usually have different sensitivities for compression, manually tuning the compression ratios is not only time-consuming, but also may lead to sub-optimal results. In the {\it structured pruning} domain, importance of a channel is typically evaluated by optimizing the reconstruction error of the feature-maps before and after pruning this channel \cite{he2017channel, luo2018autopruner, luo2017thinet}. However, these methods still require manually setting the compression ratio for each layer. Some other efforts are explored to automatically determine these compression ratios for different layers, such as using deep reinforcement learning \cite{he2018amc, lin2017runtime} or learning sparse structure networks \cite{liu2017learning}. However, at the expense of avoiding per-layer compression ratios, these methods introduce extra hyper-parameters. In addition, {\it structured pruning} methods usually rely on the fact that there are only a few thousand of channels in a neural network and the optimization problem is tractable. Therefore, it's not straight-forward for these structured pruning techniques to adapt to the {\it fine-grained} scenario, where optimization search space is so large (number of parameters) that make it impossible to solve.

In this paper, we propose NAP, a unified pruning method for both {\it fine-grained pruning} and {\it structured pruning}. Our method can automatically decide the compression ratios for different layers, and can find out unimportant parameters (channels) based on their effects on the loss function. Compared with previous methods, NAP is much easier to use (almost no hyper-parameter tuning), and shows better performance. To this end, we estimate the importance of parameters using the Taylor expansion of the loss function, similar to \cite{hassibi1993second}, and remove unimportant parameters accordingly. However, this involves computing the Hessian matrix, which is tremendously huge and intractable for modern neural networks.
Here, we first notice that the Fisher Information matrix is close to the Hessian under certain conditions. We then use a Kronecker-factored Approximate Curvature (K-FAC) \cite{martens2015optimizing} method to efficiently estimate the Fisher matrix and use it as a proxy for the Hessian. In our experiments, the overhead of estimating the Hessian approximation is small compared with the pre-train and fine-tune time. Therefore, our method is efficient to use. Importantly, this importance criterion is calibrated across layers, allowing us to automatically get the per-layer compression ratios.
We demonstrate the effectiveness of our method on several benchmark models and datasets, outperforming previous pruning methods by large margins for both {\it fine-grained pruning} and {\it structured pruning}.

\section{Related work}
\begin{figure*}[t]
\begin{center}
  \includegraphics[height=6.5cm]{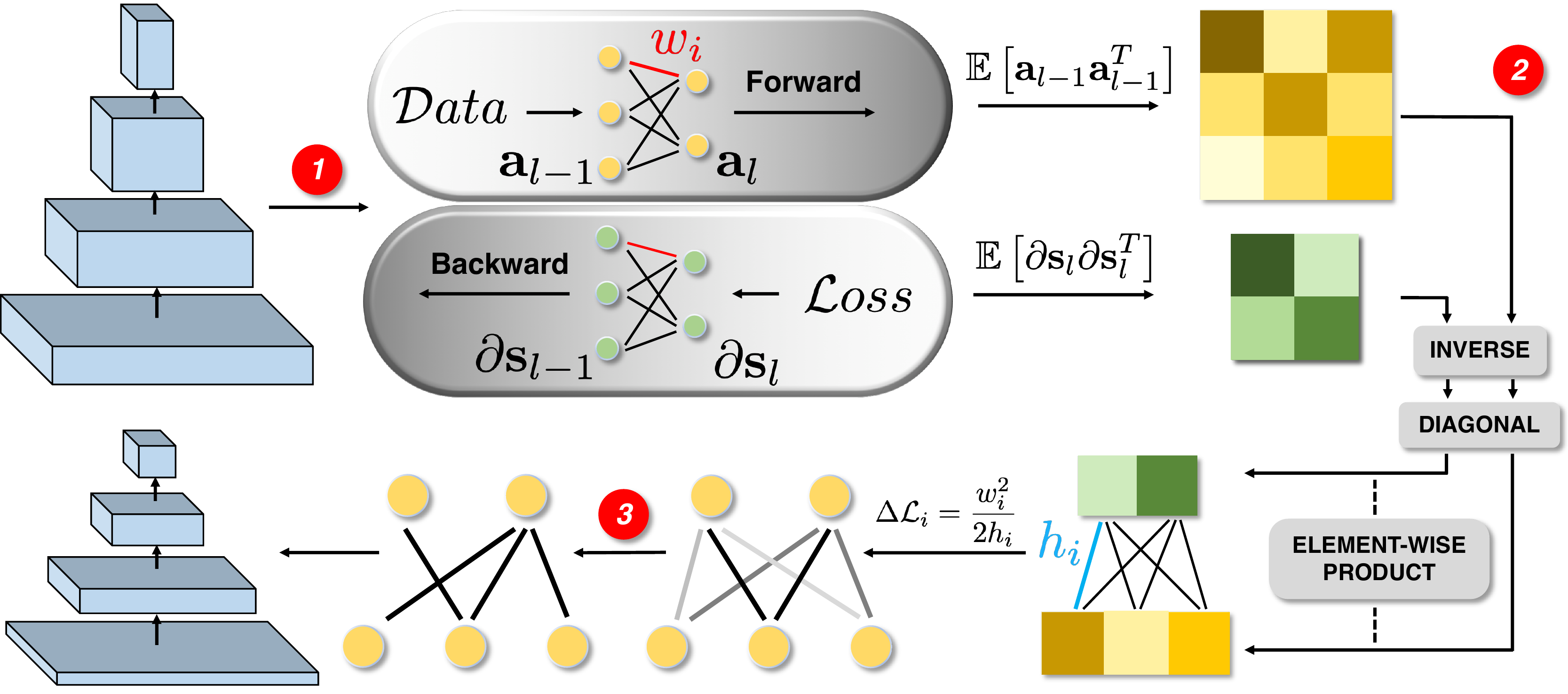}
\end{center}
\caption{NAP pipeline. \textbf{Step 1, Updating Statistics:} NAP takes $T$ steps of normal forward-backward pass to estimate statistics of activation $\mathbf{a}_{l-1}$ and derivative $\partial \mathbf{s}_{l}$ as $\mathbb{E}\left[\mathbf{a}_{l-1}\mathbf{a}_{l-1}^T \right]$ and $\mathbb{E}\left[\partial \mathbf{s}_{l}\partial\mathbf{s}_{l}^T \right]$ respectively, which are much smaller than full Hessian. \textbf{Step 2, Computing importance:} NAP then inverts and diagonalizes $\mathbb{E}\left[\mathbf{a}_{l-1}\mathbf{a}_{l-1}^T \right]$ and $\mathbb{E}\left[\partial \mathbf{s}_{l}\partial\mathbf{s}_{l}^T \right]$, do element-wise product to get $h_i$ and compute importance $\Delta \mathcal{L}_i = w_i^2 / 2h_i$ for each parameter. (See Eq. \ref{eq:solution}) \textbf{Step 3, Pruning:} NAP finally prunes $p$ fraction of parameters with lowest $\Delta \mathcal{L}_i$. These steps can be repeated multiple times. More details of NAP can be found in Section. \ref{sec:nap} and Algorithm \ref{alg:pruning}.
}
\end{figure*}

\paragraph{Model Compression}
Model compression aims to compress a large network into a smaller one while maintaining good performance. There are several popular families of compression approaches, including pruning, quantization, and low-rank approximation. Quantization \cite{courbariaux2016binarized,gong2014compressing,rastegari2016xnor,wu2016quantized,zhu2016trained} aims to use fewer bits to encode each parameter, e.g. binary neural network. Low-rank approximation \cite{denton2014exploiting,jaderberg2014speeding, lebedev2014speeding, novikov2015tensorizing} approximates network parameters by low-rank representations, saving storage and speeding up the network. Pruning, being one of the most popular methods due to its simplicity and effectiveness, aims to remove unimportant parameters from a large network. These techniques could be further integrated together and result in better compression ratios \cite{han2015deep}.

\vspace{-0.3cm}
\paragraph{Pruning}
Han \etal \cite{han2015learning} first compress modern neural networks with a magnitude-based pruning method \cite{guo2016dynamic,li2016pruning,wen2016learning}. They iteratively prune parameters with smaller value and show decent compression results. However, these {\it fine-grained pruning} methods prune individual parameters, leaving layers with irregular sparsity, which is not easy for common hardware to achieve real acceleration. To address this issue, {\it structured pruning} proposes to prune a whole channel. Since the search space is much smaller than {\it fine-grained pruning}, the {\it structured pruning} problem is typically formulated as a minimization of the reconstruction error of the feature-map before and after pruning, which can be approximated by LASSO \cite{he2017channel} or greedy algorithm \cite{luo2017thinet}. However, these methods require manually setting compression ratios for different layers, which is time-consuming and not easy to tune. Some other methods focus on automatically setting these compression ratios by, for instance, learning the importance of different channels through back-propagation \cite{liu2017learning} or deep reinforcement learning \cite{he2018amc, lin2017runtime}. However, these methods bring new hyper-parameters to tune. In addition, most of these methods can only handle either {\it fine-grained} or {\it structured} cases.

\vspace{-0.3cm}
\paragraph{OBS and K-FAC}
Although magnitude-based pruning methods are simple and show good results, they implicitly assume that parameters with smaller values are less important than others. Such an assumption doesn't generally hold as noted in \cite{hassibi1993second, lecun1990optimal}. Therefore, Optimal Brain Surgeon (OBS) \cite{hassibi1993second} proposes using Taylor expansion to estimate the importance of a parameter, and shows good performance on small networks. Since our work is inspired by OBS, we briefly review the idea here.

Given a well-trained neural network with parameters $\mathbf{W}$, the local surface of the training loss $\mathcal{L}$ can be characterized by its Taylor expansion: $\delta \mathcal{L} = \frac{1}{2} \delta \mathbf{W}^T \mathbf{H} \delta \mathbf{W}, $
where $\delta \mathbf{W}$ is a small change in the parameter values (e.g., setting some parameters to zero) and $\mathbf{H}$ is the Hessian matrix defined as $\partial^2 \mathcal{L}/\partial \mathbf{W}^2$. Note that at convergence the first-order term vanishes and the higher-order term can be neglected, and thus only the second term remains.
OBS aims to find an parameter $w_q$ such that when it is removed, the change in the training loss $\delta \mathcal{L}$ is minimized:
\begin{equation}
  \label{eq:OBS}
  \min \limits_q \left(\min \limits_{\delta \mathbf{W}} \left(\frac{1}{2} \delta \mathbf{W}^T \mathbf{H} \delta \mathbf{W} \right)\right), \quad s.t. \quad \mathbf{e}_q^T \delta \mathbf{W} + \mathbf{W}_q = 0,
\end{equation}
where $\mathbf{e}_q^T$ is the one-hot vector in the parameter space corresponding to $w_q$. The inner minimization problem can be solved by Lagrangian multipliers.
Unfortunately, since the exact solution involves the inverse of the Hessian matrix, original OBS is intractable for modern neural networks that typically contain millions of parameters. Recently, layerwise OBS \cite{dong2017learning} uses the layer-wise reconstruction loss instead of the final loss, and ends up with a smaller Hessian to evaluate. However, similar to magnitude-based methods, layerwise OBS has to tune compression ratios for different layers.

In our work, we follow the line of OBS, and extend it to modern neural network in both {\it fine-grained} and {\it structured} settings. We utilize K-FAC to approximate the Fisher matrix, which in turn approximates the exact Hessian matrix. K-FAC \cite{grosse2016kronecker, martens2015optimizing} provides an efficient way to estimate and invert an approximation of the Fisher matrix of a neural network. It first approximates the Fisher by a block diagonal matrix, and then decomposes each block by two much smaller matrices via Kronecker-product. This approximation and its variants \cite{ba2016distributed,wu2017scalable,zhang2017noisy} have shown success in the field of optimization.

\vspace{-0.2cm}
\section{NAP: Network Automatic Pruning}
\label{sec:nap}
The essence of pruning is to remove unimportant parameters, i.e. parameters that affect the model performance less but reduce the model complexity more. Therefore, we formulate pruning as an optimization problem considering both loss function and model complexity.
We first introduce this optimization problem and its ideal solution in Section \ref{sec:MLP_solution}. We then show how to make approximations to this ideal solution for practical considerations in Section \ref{sec:MLP_approximation}. Finally, we illustrate how our method can extend to channel pruning for accelerating modern neural networks in Section \ref{sec:Channel_pruning}. Our overall method is summarized in Algorithm \ref{alg:pruning}.

Through out this section, we consider a neural network with $L$ layers. The input and output for layer $l$ is $\mathbf{a}_{l-1}$ and $\mathbf{s}_l$. To indicate a parameter is retained or pruned, we introduce a mask variable $\mathbf{\Gamma}_l$ and thus the forward pass can be written as
\begin{equation}
  \mathbf{s}_l = \left(\mathbf{W}_l \odot \mathbf{\Gamma}_l\right) \mathbf{a}_{l-1}, \quad \mathbf{a}_l = \text{Relu}\left( \mathbf{s}_l \right),
\end{equation}
where $\mathbf{W}_l$ is the parameter matrix and $\odot$ denotes element-wise product.\footnote{CNN can also be written in a similar way if we expand the parameter matrix with duplicated weights.}

\subsection{Problem Formulation}
\label{sec:MLP_solution}
Our objective is to prune as much as possible while maintaining good model performance, e.g. high accuracy in classification. Similar to Minimum Description Length (MDL, \cite{rissanen1978modeling}), we formulate our objective $\Psi$ as
\begin{equation}
  \label{eq:MDL}
  \Psi = \min \limits_{\mathbf{\Theta}}  \mathcal{L} (data | \mathbf{\Theta}) + \mathcal{D}(\mathbf{\Theta}).
\end{equation}
The first term in Eq. (\ref{eq:MDL}) is the training loss $\mathcal{L}$, and the second term is a measure of the model complexity. Such measurement can take different forms, such as \textbf{1)} number of parameters, which leads to a storage-friendly small network, or \textbf{2)} number of FLOPs, which leads to a faster network. If we use number of parameters as the measurement (we'll talk about using number of FLOPs in Section \ref{sec:Channel_pruning}), the objective can then be written as
\begin{equation}
  \label{eq:objective}
  \Psi = \min \limits_{\mathbf{W}, \mathbf{\Gamma}} \left[   \mathcal{L} \left( y|x, \mathbf{W}\odot\mathbf{\Gamma} \right) + \lambda \sum_{l=1}^L  \left(\sum_{(i,j)} \gamma_l^{(i,j)}\right) \right],
\end{equation}

 where $\gamma_l^{(i,j)}$ is the mask of the $(i,j)^{th}$ parameter in the $l^{th}$ layer, and thus the second term in Eq. \ref{eq:objective} is the number of remaining parameters in this network. Therefore, solving this minimization problem will naturally prune a network in a \textit{fine-grained} manner. $\lambda$ is introduced as a relative importance which characterizes the trade-off between sparsity and model performance, and controls the final size of a compressed model.\footnote{We'll soon explain how to set the value of $\lambda$ in practical applications.}
Unfortunately, this objective cannot be directly optimized with vanilla SGD as $\gamma_l^{(i,j)}$ is constrained to be binary. Instead, we consider an easier case where we first pre-train the model to a local minimum of $\mathcal{L}$, then update one $\gamma_l^{(i,j)}$ from 1 to 0 if it decreases our objective $\Psi$.
Recall that updating the mask $\gamma_l^{(i,j)}$ is equivalent to pruning a parameter $w_l^{(i,j)}$, we can repeatedly doing so until the objective $\Psi$ converges, resulting in a smaller model.

To evaluate the effect of pruning a parameter upon $\Psi$, we use the Taylor expansion of Eq. \ref{eq:objective} and omit the first-order term similar to OBS \cite{hassibi1993second}, because we assume we start from a local minimum of $\mathcal{L}$. Suppose we want to prune one parameter $w^{(i, j)}_l$ and evaluate the change in $\Psi$, we need to solve
\begin{equation}
  \label{eq:surgeon}
   \Delta \Psi_q = \min \limits_{\mathbf{\delta W}} \left(\frac{1}{2}\mathbf{\delta W^T H \delta W} - \lambda \right), s.t. \delta w_l ^{(i,j)} + w_l^{(i,j)} = 0.
 \end{equation}
The subscript $q$ is simply an index number associated with parameter $w_l^{(i,j)}$. Essentially, we want to perturb the network in such a way ($\delta \mathbf{W}$) that updates $w_l^{(i,j)}$ from its original value to $0$ and minimize $\Delta \Psi_q$ at the same time. This optimization problem in Eq. \ref{eq:surgeon} can be solved with Lagrangian multipliers, resulting in $\Delta \Psi_q = \Delta \mathcal{L}_q - \lambda$, and
\begin{equation}
  \label{eq:solution}
  \Delta \mathcal{L}_q = \frac{1}{2}\frac{\left(w_l^{(i,j)}\right)^2}{\left[\mathbf{H}^{-1}\right]^{(q,q)}},
  \quad \mathbf{\delta W}^* = \left[-\frac{w_l^{(i,j)}}{\left[\mathbf{H}^{-1}\right]^{(q,q)}}\mathbf{H}^{-1}\right]^{(\cdot,q)}.
\end{equation}
Therefore, rather than only prune the parameter $w_l^{(i,j)}$, we also update other parameters simultaneously according to $\mathbf{\delta W}^*$, which will give us better $\Delta \Psi_q$.
In Eq. \ref{eq:solution}, superscript $(q,q)$ denotes the element at the $q^{th}$ row and $q^{th}$ column, $(\cdot,q)$ denotes the $q^{th}$ column in that matrix, and $q$ is the index of the element in $\mathbf{H}$ associated with $w_l^{(i,j)}$. The estimation of $\mathbf{H}^{-1}$ will be introduced in section \ref{sec:MLP_approximation}.

Given the solution of Eq. \ref{eq:surgeon} in Eq. \ref{eq:solution}, we can then evaluate $\Delta \Psi$ for all parameters, remove the smallest one, apply $\delta \mathbf{W}^*$ to other parameters, and repeat this procedure until the objective $\Psi$ converges. However, this is intractable for networks with millions of parameters. For practical considerations, we simultaneously remove multiple parameters with lower $\Delta \Psi_q$, i.e. we first evaluate $\Delta \Psi_q$ for all parameters, remove parameters that have $\Delta \Psi_q < 0$, and update others with $\delta \mathbf{W^*}$ of those removed parameters.

Recall that our ultimate goal is to achieve a small model. To do so, one can set an appropriate $\lambda$ value in the very beginning, repeat the aforementioned procedure several iterations\footnote{Such iterative manner is commonly adopted in pruning methods.}, until $\Psi$ doesn't decrease. However, the value of $\lambda$ is critical for finding a sweet-point between model size and model performance, and it's generally not easy to set an appropriate $\lambda$ beforehand. Therefore, we instead dynamically adjust the value of $\lambda$ as pruning proceeds. At each iteration, we set $\lambda$ such that the aforementioned pruning operation will give us a smaller model with $(1-p)$ times of original size. This is equivalent to setting $\lambda$ to the $p^{th}$ percentile of $\Delta \mathcal{L}_q$. By iteratively doing this, we can monitor the model size and model performance, and thus easily decide when to stop pruning. More importantly, this also avoids tuning hyper-parameter $\lambda$, and only leave us with $p$. As we'll show in our ablation study, our method is robust to the value of $p$, and thus it is easy to apply.

After pruning, we fine-tune the remaining parameters using SGD to get better model performance,

\subsection{Approximating Hessian using Fisher}
\label{sec:MLP_approximation}
Performing the pruning operation as in Eq. (\ref{eq:solution}) involves estimating and inverting the Hessian matrix, which is intractable for modern neural networks that contain millions of parameters. Therefore, we propose an approximation of the Hessian to efficiently calculate Eq. (\ref{eq:solution}). We first employ the Fisher Information matrix as an approximation of the Hessian and further use a Kronecker-factored Approximate Curvature (K-FAC) method to approximate the Fisher matrix. The first approximation comes from the fact that if the training objective is the negative log-likelihood, the Hessian matrix and Fisher matrix are the expectations of second-order derivatives under the data distribution and model distribution respectively. Since modern neural networks usually have strong model capacities, we expect those two distributions are close for a well-trained model. The second approximation (K-FAC) is demonstrated to be effective in optimization tasks \cite{grosse2016kronecker, martens2015optimizing}, and will further help us to calculate Eq. (\ref{eq:solution}) efficiently.

Given a neural network with stacked fully-connected layers\footnote{Details for convolutional layers can be found in \cite{grosse2016kronecker}.}, the Fisher Information matrix is defined as
\begin{equation}
\vspace{-0.2cm}
\mathbf{F} = \mathbb{E} \left[ \left(\nabla_{\overrightarrow{\mathbf{W}}} \mathcal{L}\right) \left(\nabla_{\overrightarrow{\mathbf{W}}} \mathcal{L}\right)^T \right],
\end{equation}
where $\overrightarrow{\mathbf{W}}$ is the vectorization of all parameters $\mathbf{W}$ in this network.\footnote{Unless specified otherwise, the expectation is taken with respect to the model distribution.} Thus, $\mathbf{F}$ has same dimension as number of parameters. The K-FAC method approximates this Fisher matrix $\mathbf{F}$ using a block-diagonal matrix, and estimates each block by the Kronecker-product of two much smaller matrices, which are the second-order statistics of inputs and derivatives. To illustrate the idea of K-FAC, we first re-write $\mathbf{F}$ in a block-wise manner by partitioning rows and columns of $\mathbf{F}$ if they correspond to parameters within the same layer. The $(i,j)$ block is then
\vspace{-0.2cm}
\begin{equation}
\mathbf{F}_{ij} = \mathbb{E} \left[ \left( \nabla_{\overrightarrow{\mathbf{W}_i}} \mathcal{L} \right) \left( \nabla_{\overrightarrow{\mathbf{W}_j}} \mathcal{L}\right)^T \right].
\end{equation}
As noted by \cite{martens2015optimizing}, the off-diagonal term $\mathbf{F}_{ij}$ is generally much smaller than diagonal term $\mathbf{F}_{ii}$.
Therefore, $\mathbf{F}$ can be approximated by a block-diagonal matrix and the $l^{th}$ diagonal block is  $\mathbf{F}_{ll}$.

Using back-propagation, the gradients of layer $l$ can be expressed as $\nabla_{\overrightarrow{\mathbf{W}_l}}\mathcal{L} = \text{vec} \left\{  \left(\nabla_{\mathbf{s}_l} \mathcal{L}\right) \left(\mathbf{a}_{l-1}\right)^T \right\}$\footnote{$\text{vec}\{\}$ denotes vectorization.}. The $l^{th}$ diagonal block $\mathbf{F}_{ll}$ can then be written as
\begin{align}
  \label{eq:kfac}
  \mathbf{F}_{ll} &= \mathbb{E}\left[ \text{vec} \left\{ \left(\nabla_{\mathbf{s}_l} \mathcal{L}\right) \left(\mathbf{a}_{l-1}\right)^T \right\} \text{vec} \left\{ \left(\nabla_{\mathbf{s}_l} \mathcal{L}\right) \left(\mathbf{a}_{l-1}\right)^T \right\}^T \right] \\
  &= \mathbb{E}\left[ \mathbf{a}_{l-1}\mathbf{a}_{l-1}^T \otimes (\nabla_{\mathbf{s}_l} \mathcal{L})(\nabla_{\mathbf{s}_l} \mathcal{L})^T \right]\\
  &\approx \mathbb{E}\left[ \mathbf{a}_{l-1}\mathbf{a}_{l-1}^T\right] \otimes \mathbb{E}\left[(\nabla_{\mathbf{s}_l} \mathcal{L})(\nabla_{\mathbf{s}_l} \mathcal{L})^T \right],
\end{align}
where $\otimes$ denotes Kronecker-product. The second equality  comes from the property of the Kronecker-product\footnote{$vec\left\{\mathbf{uv}^T\right\} = \mathbf{v} \otimes \mathbf{u}$,  \quad $(\mathbf{A}\otimes \mathbf{B})(\mathbf{C}\otimes \mathbf{D}) = (\mathbf{A}\otimes \mathbf{C})(\mathbf{B}\otimes \mathbf{D})$.}.
The last approximation is to further accelerate the computation, and has shown to be effective in optimization domain \cite{martens2015optimizing}.
Therefore, $\mathbf{F}$ (and thus  $\mathbf{H}$) can be approximated by several much smaller matrices
\vspace{-0.2cm}
\begin{gather}
   \mathbf{A}_{l-1} = \mathbb{E}\left[ \mathbf{a}_{l-1}\mathbf{a}_{l-1}^T\right], \quad \mathbf{DS}_l = \mathbb{E}\left[(\nabla_{\mathbf{s}_l} \mathcal{L})(\nabla_{\mathbf{s}_l} \mathcal{L})^T \right],\\
   \label{eq:kfacstats}
   \mathbf{F}_{ll} = \mathbf{A}_{l-1} \otimes \mathbf{DS}_l.
   \vspace{-0.2cm}
 \end{gather}
  For a typical modern neural network such as AlexNet, the original Hessian matrix is a $61\text{M} \times 61\text{M}$ matrix, while $\mathbf{A}_{l-1}$ and $\mathbf{DS}_l$ of the largest fully-connected layer have sizes of only $9126 \times 9126$ and $4096 \times 4096$ respectively, and we only need to estimate one $\mathbf{A}_{l-1}$ and $\mathbf{DS}_l$ for each layer. We use exponential moving average to estimate the expectation in $\mathbf{A}_{l-1}$ and $\mathbf{DS}_l$, with a decay factor of 0.95 in all experiments. This adds only a small overhead to the normal forward-backward pass. Furthermore, the inverse $\mathbf{F}^{-1}$ and matrix-vector product $\mathbf{F}^{-1}\mathbf{h}$ can also be efficiently calculated by leveraging the block diagnoal structure and property of Kronecker-product\footnote{$(\mathbf{A} \otimes \mathbf{B})^{-1} = \mathbf{A}^{-1} \otimes \mathbf{B}^{-1}$, \quad $(\mathbf{A} \otimes \mathbf{B})vec\{\mathbf{X}\} = vec\{\mathbf{BXA}^T\}$}.

With Eq. \ref{eq:kfacstats}, now we can compute $\Delta \mathcal{L}_q$ in Eq. \ref{eq:solution} efficiently. Ideally, $\Delta \mathcal{L}_q$ should capture the effect of pruning upon training loss, and thus should be calibrated across all layers. For example, given one neuron, the sum of $\Delta \mathcal{L}_q$ for all incoming parameters in the previous layer should be close to the sum for all outgoing parameters, because deleting all incoming parameters is the same as deleting all outgoing parameters. However, we empirically find these sums sometimes differ by a scale factor from layer to layer. This might be due to the block diagonal approximation decorrelating some cross-layer influences, which is an interesting topic for future work. In this paper, we employ a simple yet effective strategy:
after calculating $\Delta \mathcal{L}_q$, we normalize it within the same layer: $\tilde{\Delta \mathcal{L}_q} = \Delta \mathcal{L}_q / \sum_{q \in \text{layer } l}\Delta \mathcal{L}_q,$
and change our importance measure $\Delta \Psi_q$ accordingly. $\delta \mathbf{W^*}$ and other steps remain unchanged.

To summarize, our method first takes $T$ steps of normal forward-backward pass on training dataset, and estimates statistics $\mathbf{A}_{l-1}$ and $\mathbf{DS}_l$ for each layer. It then combines Eq. \ref{eq:solution} and Eq. \ref{eq:kfacstats} to compute $\Delta \mathcal{L}_q$ ($\tilde{\Delta \mathcal{L}_q}$), removes smallest $p$ fraction parameters, updates other parameters accordingly, and then does another iteration. We show our algorithm in Algorithm \ref{alg:pruning}.

\begin{algorithm}[t]
\caption{NAP: Network Automatic Pruning}
\label{alg:pruning}
\begin{algorithmic}[1]
\REQUIRE  $\mathbf{W}$, $\mathbf{\Gamma}$, %
Pruning fraction $p$, Number of steps $T$ for estimating $\mathbf{A}_{l-1}$ and $\mathbf{DS}_l$, %
\pretrain
\STATE Pre-train the network

\pruning
  \FOR{$t=0, \cdots, T$}
    \STATE Update $\mathbf{A}_{l-1} \gets  0.95\times\mathbf{A}_{l-1} + 0.05\times\mathbf{a}_{l-1}\mathbf{a}^T_{l-1}$

    \STATE Update $\mathbf{DS}_l \gets 0.95\times\mathbf{DS}_l + 0.05\times(\nabla_{\mathbf{s}_l} \mathcal{L})(\nabla_{\mathbf{s}_l} \mathcal{L})^T$
  \ENDFOR

  \STATE Compute Fisher matrix $\mathbf{F}$ by Eq. (\ref{eq:kfacstats}).
  \STATE Compute importance measure $\Delta \mathcal{L}_q$ for each parameter by Eq. (\ref{eq:solution}), and normalize within each layer to get $\tilde{\Delta \mathcal{L}_q}$.
  \STATE Compute $p^{th}$ percentile of $\tilde{\Delta \mathcal{L}_q}$ from all layers as $\lambda$.
  \STATE Update mask $\gamma_l^{(i,j)}$ to $0$ if its corresponding $\tilde{\Delta \mathcal{L}_q}$ is smaller than $\lambda$.
  \STATE Compute $\delta \mathbf{W}^*$ by Eq. (\ref{eq:solution}) and update $\mathbf{W} \gets \mathbf{W}+\delta\mathbf{W}^*$.

\retrain
\STATE Fine-tune the network using SGD.

\end{algorithmic}

\end{algorithm}

\subsection{Channel-wise Pruning}
\label{sec:Channel_pruning}
In Eq. \ref{eq:objective}, we use number of parameters as the model complexity measurement. This leads us to a small model with irregular sparsity, i.e. some individual parameters are pruned away. However, this doesn't imply faster inference in reality. First, a small number of parameters doesn't necessarily mean a small number of FLOPs, as a fully connected layer usually has a much larger number of parameters than a convolution layer, but a much fewer number of FLOPs. Second, irregular sparsity typically needs specialized hardware/software design for acceleration. To achieve faster speed during inference, we now extend our method to consider number of FLOPs and channel-wise pruning.

Channel-wise pruning prunes a whole channel from a convolution layer, and thus directly achieves higher forward and backward speed. To estimate the importance of a channel $C$ in this network, we need to evaluate $\Delta \Psi$ in Eq. \ref{eq:objective}, which includes the effect on loss $\Delta \mathcal{L}_C$ and the effect on model complexity. Note that pruning a channel means pruning all the parameters in that channel. Therefore, the change in training loss $\Delta \mathcal{L}_C$ can be approximated by the sum of $\tilde{\Delta \mathcal{L}_q}$ of all parameters in this channel. The second term in Eq. \ref{eq:objective} is how much we reduce the model complexity. Here, we use number of FLOPs as a measurement, and the second term becomes how many FLOPs are reduced by pruning this channel. Similar to the {\it fine-grained pruning} setting, we set $\lambda$ to the $p^{th}$ percentile of $\Delta \mathcal{L}_C$/$\Delta \text{FLOPs}$ of all channels in this network, and remove those channels smaller to $\lambda$. This give us a 'thinner' network with fewer FLOPs.

\section{Experiments}
As illustrated in Section \ref{sec:nap}, NAP can handle both \textit{fine-grained pruning} and \textit{structured pruning}. Therefore, we demonstrate the effectiveness of NAP in both of these two settings. We'll first show in Section \ref{sec:structured} the results of \textit{structured pruning},
and then show our method can also push the limit of \textit{fine-grained pruning} in Section \ref{sec:finegrained}. For \textit{structured pruning}, we use speed-up ratio based on FLOPs as our metric and also report the real inference acceleration tested on GPU. For \textit{fine-grained pruning}, we use compression ratio based on number of parameters.

To show the generality of our method, we conduct experiments on six popular deep network architectures, from shallow to deep, and a number of benchmark datasets. This  includes LeNet-300-100 and LeNet-5 \cite{lecun1998gradient} on MNIST, CifarNet \cite{krizhevsky2009learning} on Cifar-10, AlexNet \cite{krizhevsky2012imagenet},  VGG16\cite{simonyan2014very} and ResNet-50\cite{he2016deep} on Imagenet ILSVRC-2012.
The aforementioned architectures contain both fully-connected layers and convolution layers, with different sizes (from 267K to 138M parameters) and different depth (from 3 to 50).
In addition, we also test the generalization of our compressed model on detection benchmarks.
More implementation details and analysis of NAP can be found in the appendix. %

\begin{table}[t]
  \centering
  \small
  \begin{tabular}{c|c|cccc}
    \hline
    \hline
    Network & Method & $\Delta$ Top1 & $\Delta$ Top5 & Speed up (x)\\
    \hline
    \multirow{9}{*}{VGG16}
    &TE\cite{molchanov2016pruning} & - & 4.8 & 3.9 \\
    &FP\cite{li2016pruning} & - & 8.6 & $\approx 4$\\
    &Asym\cite{zhang2016accelerating} & - & 3.8 & $\approx 4$\\
    &SSS\cite{huang2018data} & 3.9 & 2.6 & 4.0\\
    &CP\cite{he2017channel} & 2.7 & 1.7 & 4.5 \\
    &ThiNet\cite{luo2017thinet} & 2.4 & 1.2 & 4.5\\
    &RNP\cite{lin2017runtime} & - & 3.6 & $\approx 5$ \\
    &AMC\cite{he2018amc} & - & 1.4 & $\approx 5$\\
    \cline{2-5}

    &\textbf{NAP} & \textbf{2.3} & \textbf{1.2} & \textbf{5.4}\\
    \hline
    \multirow{4}{*}{ResNet50}
    &SSS\cite{huang2018data} & 4.3 & 2.1 & 1.8 \\
    &CP\cite{he2017channel} & 3.3 & 1.4 & $\approx 2$\\
    &ThiNet\cite{luo2017thinet} & 4.3 & 2.2 & 2.3\\

    \cline{2-5}
    &\textbf{NAP} & \textbf{2.0} & \textbf{1.1} & \textbf{2.3} \\
    \hline
    \hline

  \end{tabular}
  \caption{Increase of classification error on ImageNet with channel-wise pruning (lower the better). As a reference, our pre-trained VGG16 has accuracy of 71.0\% / 89.8\% and ResNet50 of 75.0\% / 92.2\% (Top1 / Top5, single-crop).}
  \vspace{-0.5cm}
  \label{table:channel}
\end{table}

\subsection{Structured Pruning for Model Acceleration}
\label{sec:structured}
Following our discussion in Section. \ref{sec:Channel_pruning}, we apply NAP to conduct channel-wise pruning. This is an important application of compression methods, as channel-wise pruning can reduce FLOPs and directly accelerate models on common hardware, e.g. GPU.
Here, we follow previous work and demonstrate the effectiveness of our method on ImageNet dataset with VGG16 and ResNet50. We compare ours performance with previous channel (filter) pruning methods in Table \ref{table:channel}. These methods are sorted by their FLOPs speed-up ratio, and we use notation $\approx$ if the exact FLOPs number is not reported in the original paper.

\vspace{-0.3cm}
\subsubsection{VGG16 Acceleration}
VGG16 has computation intensive convolutional layers, e.g. big feature-map size and big channel numbers. Therefore, it potentially has big redundancies in its channels and acceleration is possible.
To compress the network, we iteratively update statistics for 0.2 epoch and prune 1\% of remaining channels until a favorable model size is reached. Note that more steps for updating statistics and smaller pruning ratio are possible and may help the final performance (see Appendix \ref{app:hyper}), but will consume more time for pruning. Similar to ThiNet \cite{luo2017thinet}, we fine-tune the final compressed model with learning rate varying from 1e-3 to 1e-4, and stop at 30 epochs.

As shown in Table. \ref{table:channel}, NAP outperforms all those recent channel-pruning methods, with highest speed-up ratio and smallest performance drop. In addition, we notice that: \textbf{1)} NAP only requires the pruning ratio $p$ and updating steps $T$ (or fine-tuning steps) in one pruning iteration. As we'll show in Section \ref{sec:ablation}, NAP is robust to the values of $p$ and $T$, and thus our method is easy to use in real applications. On the contrary, in addition to $p$ and $T$, most of the previous methods in Table \ref{table:channel} either require manually tuned per-layer compression ratios \cite{li2016pruning, zhang2016accelerating, he2017channel, luo2017thinet, wang2017structured}, or introduce new hyper-parameters \cite{lin2017runtime, he2018amc}. \textbf{2)} Similar to our method, TE \cite{molchanov2016pruning} approximates the effect of pruning one parameter using Taylor expansion. It uses the variance of gradient rather than hessian to evaluate the Taylor expansion. However, this method performs not as well as ours. This empirically suggests that, evaluating the second order information is necessary for the Taylor expansion to estimate the parameters' importance. \textbf{3)} We also test on GPU the absolute acceleration of inference time. After compression, our model achieves \textbf{2.74} times faster than the original VGG16.\footnote{Testing with 1080 Ti, Tensorflow 1.8, CUDA 9.2, cuDNN 7.1 and batch size 64.}

\begin{table}[t]
  \centering
  \begin{tabular}{c|ccc}
    \hline
    \hline

    Method & Baseline & Compressed & Speed up (x)\\
    \hline
    CP(2x)\cite{he2017channel} & 68.7 & 68.3 & $\approx 2$\\
    CP(4x)\cite{he2017channel} & 68.7 & 66.9 & $\approx 4$\\
    Asym\cite{zhang2016accelerating}\tablefootnote{Reported in \cite{he2018amc}} & 68.7 & 67.8 & $\approx 4$\\
    AMC\cite{he2018amc} & 68.7 & 68.8 & $\approx 4$\\

    \hline
    \textbf{NAP} & 70.0 & 68.8 & \textbf{5.4} \\
    \hline
    \hline
  \end{tabular}
  \caption{Faster-RCNN detection result on PASCAL VOC 2007, using compressed VGG16 as the backbone network. We report the mAP (\%) for both baseline model and compressed model. Speed-up is computed based on the flops of VGG backbone, not the whole detector.}
  \vspace{-0.5cm}
  \label{table:detection}
\end{table}

\subsubsection{ResNet50 Acceleration}
\vspace{-0.1cm}
ResNet has fewer FLOPs yet higher performance than VGG16. Therefore, it has less redundancies in its channels and more challenging to prune. In addition, the residue structure typically requires further modifications of pruning methods \cite{he2017channel, liu2017learning}, because the {\it shortcut} layers and {\it branch\_c} layers need to have same channel numbers. To apply our method on ResNet50, we treat a channel in a {\it shortcut} layer and its corresponding channels in other {\it branch\_c} layers of the same stage as one unity (a virtual channel). Thus, pruning one such unity effectively reduces 1 channel from the shorcut path, and will maintain the residue structure. Recall that we compute the importance of a channel by $\Delta \mathcal{L}_C$/$\Delta \text{FLOPs}$ (Section. \ref{sec:Channel_pruning}), we can compute the importance of such unity by aggregating from all associated channels, i.e. the sum of $\Delta \mathcal{L}_C$ of a channel in the {\it shortcut} layer and its corresponding channels in {\it branch\_c} layers, divided by the sum of $\Delta \text{FLOPs}$ after removing these channels. After computing the importance, this unity also compares its importance with other channels, and we simply removes $k$ less important channels in this network.

The compression result is shown in Table. \ref{table:channel}. Compared with previous methods, we achieve smaller performance drop under similar speed-up ratio. We notice that previous methods \cite{he2017channel, luo2017thinet} usually manually define the structure of the compressed network, e.g. channels in the shortcut path  remain unchanged and blocks closer to max-pooling are pruned less. However, such a structure is not easy to tune and may not be optimal for model compression, and thus limit their performance. Furthermore, these methods also have difficulties to achieve real acceleration on GPU (\cite{he2017channel, liu2017learning} relies on sampling layers which make the compressed ResNet even slower than the orignal one.) Since our method directly prunes unimportant shortcut channels as well as other channels, we don't need to do any customization for acceleration: when testing inference time on GPU, our compressed ResNet50 can run more than \textbf{1.4} times faster than the original one.

\subsubsection{Generalization to Object Detection}

We further test the generalization of our pruned model on detection task. After compressing the VGG16 on ImageNet, we use it as the backbone network and train Faster-RCNN on PASCAL VOC 2007 dataset. As shown in Table. \ref{table:detection}, our model can match the absolute performance of previous best model, with higher speed-up ratio. Compared with our uncompressed baseline model, we have only 1.2\% performance drop under 5.4 times backbone network acceleration. This suggests that our model indeed preserves enough model capacity and useful features, and can generalize to other tasks.

\subsection{Fine-grained Pruning for Model Compression}
\label{sec:finegrained}
Our method can also be applied to fine-grained settings. Here, we compare our method with recent fine-grained pruning methods, including 1) Randomly Pruning \cite{dong2017learning}, 2) OBD \cite{lecun1990optimal}, 3) LWC \cite{han2015learning}, 4) DNS \cite{guo2016dynamic}, 5) L-OBS \cite{dong2017learning} and 6) AMC \cite{he2018amc}. We report the change in top-1 error before and after pruning (except for ResNet50 we report top-5, since our baselines \cite{dong2017learning, han2015learning, he2018amc} only show results for top-5 error). Across different datasets and architectures, our method shows better compression results.

\begin{table}[t]

  \centering
  \begin{tabular}{c|lcc}
    \hline
    \hline
    Architecture & Method & $\Delta$ Error & CR\\
    \hline
    \multirow{6}{*}{LeNet-300-100}
    & Random \cite{dong2017learning} & 0.49\% & 12.5 \\
    & OBD \cite{lecun1990optimal} & 0.20\% & 12.5 \\
    & LWC \cite{han2015learning} & -0.05\% & 12.5 \\
    & DNS \cite{guo2016dynamic} & \textbf{-0.29\%} & 55.6 \\
    & L-OBS \cite{dong2017learning} & 0.20\% & 66.7 \\
    \cline{2-4}
    & \textbf{NAP} & 0.08\% & \textbf{77.0}\\
    \hline
    \multirow{5}{*}{LeNet-5}
    & OBD \cite{lecun1990optimal} & 1.38\% & 12.5 \\
    & LWC \cite{han2015learning} & \textbf{-0.03\%} & 12.5 \\
    & DNS \cite{guo2016dynamic} & 0.00\% & 111 \\
    & L-OBS \cite{dong2017learning} & 0.39\% & 111 \\
    \cline{2-4}
    & \textbf{NAP} & 0.04\% & \textbf{200} \\
    \hline
    \multirow{3}{*}{CifarNet}
    & LWC \cite{han2015learning} & 0.79\% & 11.1 \\
    & L-OBS \cite{dong2017learning} & 0.19\% & 11.1 \\
    \cline{2-4}
    & \textbf{NAP} & \textbf{0.17\%} & \textbf{15.6} \\
    \hline
    \hline
  \end{tabular}
  \caption{MNIST and Cifar-10 Results. CR means Compression Ratio, higher the better. $\Delta$ Error is the increase of classification error, lower the better.}
  \label{table:mnist_result}
  \vspace{-0.3cm}
\end{table}

\vspace{-0.3cm}
\paragraph{MNIST:}
We first conduct experiments on the MNIST dataset with LeNet-300-100 and LeNet-5. LeNet-300-100 has 2 fully-connected layers, with 300 and 100 hidden units respectively. LeNet-5 is a CNN with 2 convolutional layers, followed by 2 fully-connected layers. %
Table \ref{table:mnist_result} shows that we can compress LeNet-300-100 to 1.3\% of original size with almost no loss in performance. Similarly for LeNet-5, we can compress to 0.5\%, which is much smaller than previous best result.

\paragraph{Cifar-10:}
We conduct experiments on Cifar-10 image classification benchmark with Cifar-Net architecture. Cifar-Net is a variant of AlexNet, containing 3 convolutional layers and 2 fully-connected layers.
Following previous work \cite{dong2017learning}, we first pre-train the network to achieve 18.43\% error rate on the testing set. After pruning, our method can compress the original model into a 16 times smaller one with negligible accuracy drop.

\begin{table}[t]
  \centering
  \begin{tabular}{c|ccc}
    \hline
    \hline
    Architecture & Method & $\Delta$ Error & CR(x)\\
    \hline
    \multirow{4}{*}{AlexNet}
    & LWC \cite{han2015learning} & -0.01\% & 9.1 \\
    & L-OBS \cite{dong2017learning} & -0.19\% & 9.1 \\
    & DNS \cite{guo2016dynamic} & \textbf{-0.33\%} & 18 \\
    \cline{2-4}
    & \textbf{NAP} & -0.03\% & \textbf{25} \\
    \hline
    \multirow{3}{*}{VGG16}
    & LWC \cite{han2015learning} & -0.06\% & 13 \\
    & L-OBS \cite{dong2017learning} & 0.36\% & 13 \\
    \cline{2-4}
    & \textbf{NAP} & \textbf{-0.17}\% & \textbf{25} \\
    \hline
    \multirow{4}{*}{ResNet50}
    & L-OBS \cite{dong2017learning} & 2.2\% & 2.0 \\
    & LWC \cite{han2015learning} & \textbf{-0.1}\% & 2.7 \\
    & AMC \cite{he2018amc} & -0.03\% & 5.0 \\
    \cline{2-4}
    & \textbf{NAP} & 0.05\% & \textbf{6.7}\\
    \hline
    \hline
  \end{tabular}
  \caption{ImageNet Results. CR means Compression Ratio, higher the better. $\Delta$ Error is the increase of classification error, lower the better.}
  \label{table:imagenet_result}
\end{table}

\paragraph{ImageNet:}
To demonstrate our method's effectiveness on larger models and datasets, we prune AlexNet on ImageNet ILSVRC-2012.
As shown in Table. \ref{table:imagenet_result}, we achieve 25 times compression of its original size. Different from DNS \cite{guo2016dynamic} which allows pruned parameter to revive in order to recover wrong pruning decisions, our method only removes parameter, yet achieve better performance. We think this empirically suggest that our method propose a better criterion to decide unimportant parameters, and make fewer wrong pruning decisions.
Also notice that previous work \cite{guo2016dynamic, han2015learning} find out it's necessary to fix the convolutional layers' weights when pruning and retraining the fully-connected layer (and vice versa), in order to recover the performance drops from pruning. However, we don't observe such difficulty in our experiments, as we simply prune and retrain all layers simultaneously. This also indicates that our pruning operation only brings recoverable errors and still preserves the model capacity.

\begin{table}[t]
  \centering
  \begin{tabular}{cc|ccc}
    \hline
    \hline
    Ratio (\%) & Epoch & $\Delta$ Top1 & $\Delta$ Top5 & Speed up (x)\\
    \hline
    0.5 & 0.1 & 2.4 & 1.4 & 5.4\\
    1 & 0.2 & \textbf{2.3} & \textbf{1.2} & \textbf{5.4} \\
    2 & 0.4 & 2.6 & 1.3 & 5.4\\
    5 & 1 & 2.5 & 1.3 & 5.3 \\
    10 & 2 & 2.6 & 1.3 & 4.9 \\

    \hline
    \hline
  \end{tabular}
  \caption{VGG16 channel-wise pruning results with different hyper-parameters. The first column is pruning percentage $p$. The second column is number of update steps $T$. The results show NAP is robust to hyper-parameters' values.}
  \vspace{-0.4cm}
  \label{table:hyperparameter}
\end{table}

We then apply our method to a modern neural network architecture, VGG16.
Despite its large number of parameters and deep depth, we find no difficulties to prune VGG16 using our method. We use the same pruning set-up as in AlexNet experiment, and achieve the best result as shown in Table. \ref{table:imagenet_result}.

However, both AlexNet and VGG16 have large fully-connected layers, which contribute most of the network parameters. It's unclear whether the success of our method can generalize to architectures mainly consisted of convolutional layers. Therefore, we further conduct experiment on ResNet50, which has 49 convolutional layers, and one relatively small fully-connected layer. Due to its large number of layers, the search space of per-layer compression ratios is exponentially larger than that of AlexNet and VGG16. Therefore, conventional methods would have difficulties to tune and set the compression ratio for each of those 16 layers. This is supported by the fact that automatic methods like AMC and ours achieve much better results than manually tuned ones like L-OBS and LWC, as shown in Table \ref{table:imagenet_result}. Furthermore, the convolutional layers usually are much less redundant than fully-connected layers, and thus a wrong choice of parameters to prune may lead to severe and unrecoverable performance drops. This explains why our method outperform AMC: though benefiting from reinforcement learning to predict per-layer compression ratios, AMC defines unimportant parameters as those having lower magnitude. On the contrary, our method use a theoretically sound criterion, and thus resulting in better performance.

\subsection{Ablation Study}
\label{sec:ablation}
As shown in Table. \ref{table:hyperparameter}, we choose different pruning ratio $p$ and number of update steps $T$ to perform pruning on VGG16. The essential idea is to keep the total running time (epochs) for pruning roughly the same, while changing different $p$ and $T$. From the table we can observe that NAP is very robust to the hyper-parameter values. As long as hyper-parameter values remain in a reasonable range, NAP can achieve decent results automatically. We'll provide more ablation study in the supplementary materials.

\vspace{-0.2cm}
\section{Conclusion}

In this paper we have proposed Network Automatic Pruning method. It can be run
in an almost hyper-parameter free manner, and show descent compression results.
We've test its effectiveness on both {\it fine-grained pruning} and {\it
structured pruning} settings and show strong performance.

{\small
\bibliographystyle{ieee}
\bibliography{our_ref_clean}
}

\clearpage
\appendix

\section{Does NAP prune similar parameters as magnitude-based pruning?}

Different from conventional pruning methods, NAP select unimportant parameters based on their effects upon the final loss function. Here, we'll investigate what parameters NAP prune within a layer. Since both NAP and magnitude-based pruning can achieve good compression results, it's intriguing to explore if NAP prunes similar parameters as magnitude-based pruning, or they prune totally different parameters. We conduct our method in the {\it fine-grained} setting, which provides us more fine-grained information.

\begin{figure}[htb]
    \centering
        \centering
        \includegraphics[height=1.8in]{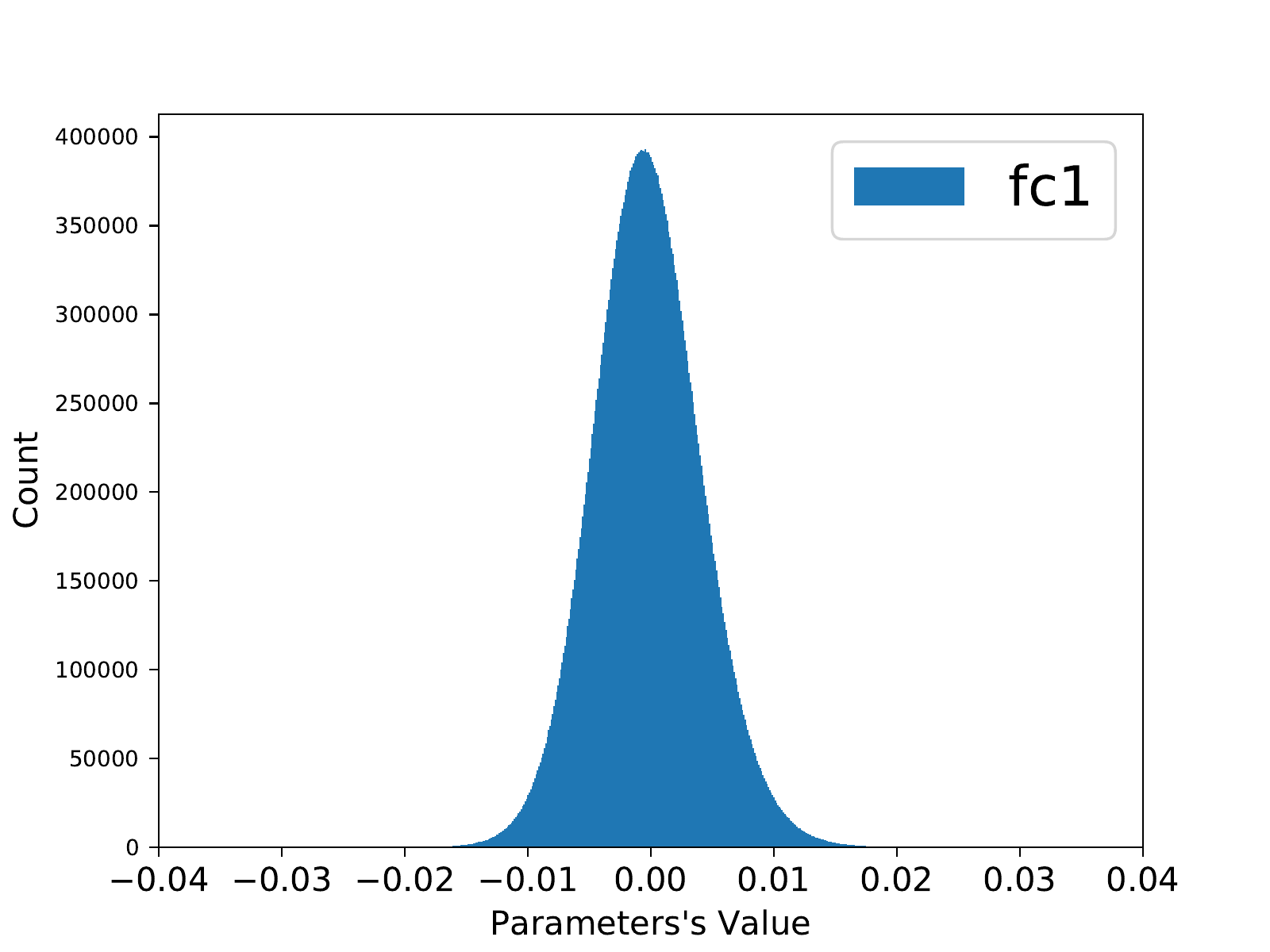}
        \caption{Weight distribution before pruning (First FC in AlexNet).}
        \label{fig:wd_before}
\end{figure}

\begin{figure}[htb]
        \centering
        \includegraphics[height=1.8in]{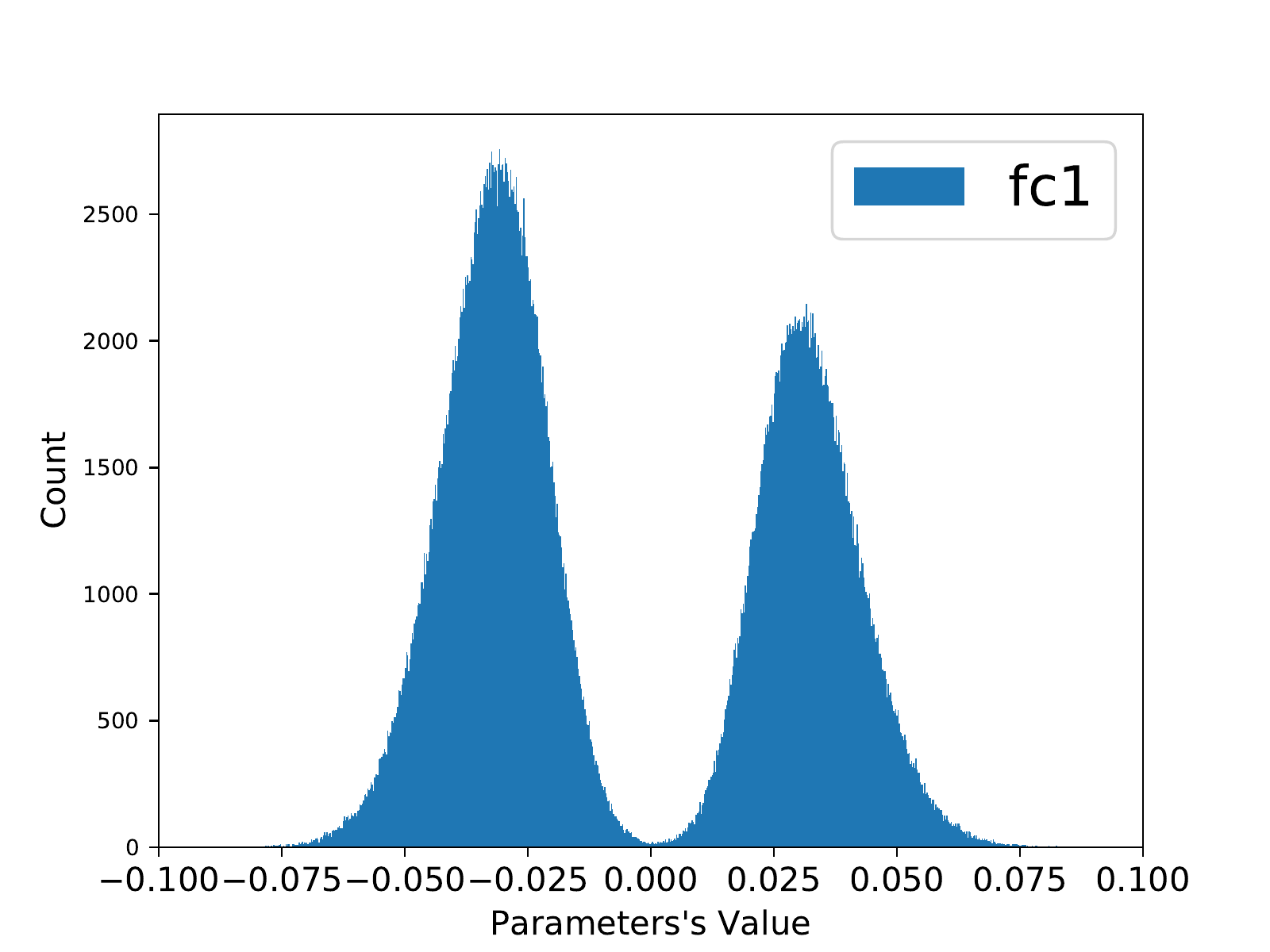}
        \caption{Weight distribution after pruning (First FC in AlexNet).}
        \label{fig:wd_after}

\end{figure}

Figure \ref{fig:wd_before} and Figure \ref{fig:wd_after} show the distribution of parameters' magnitudes, coming from the first fully-connected layer of AlexNet. Before pruning, weight distribution is peaked at 0, and drops quickly as the absolute value increasing. This is very much like a gaussian distribution as what the weights were initialized from. Figure \ref{fig:wd_after} shows the distribution after pruning using our method. It is obvious to see that parameters with magnitudes close to 0 (center region) are pruned away, indicating that the parameters regarded as having small impacts upon the loss by our method also usually have small magnitudes. Another interesting observation is that after pruning, the magnitudes of remaining parameters are larger than before, i.e. parameters' magnitudes are around 0.03 where they are typically smaller than 0.01 before pruning. This may provide some intuitions for future work on designing new initialization distributions.

\begin{figure*}[htb]
    \centering
    \begin{subfigure}[t]{0.48\textwidth}
        \centering
        \includegraphics[height=1.8in]{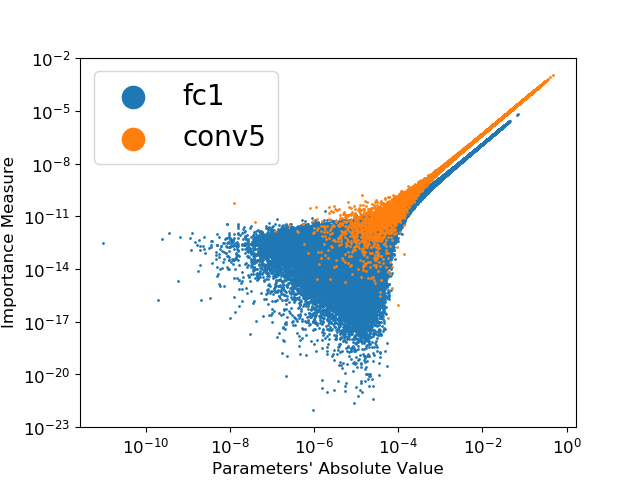}
        \caption{100\% prune to 50\%}
        \label{fig:100to50}
    \end{subfigure}%
    ~
    \begin{subfigure}[t]{0.48\textwidth}
        \centering
        \includegraphics[height=1.8in]{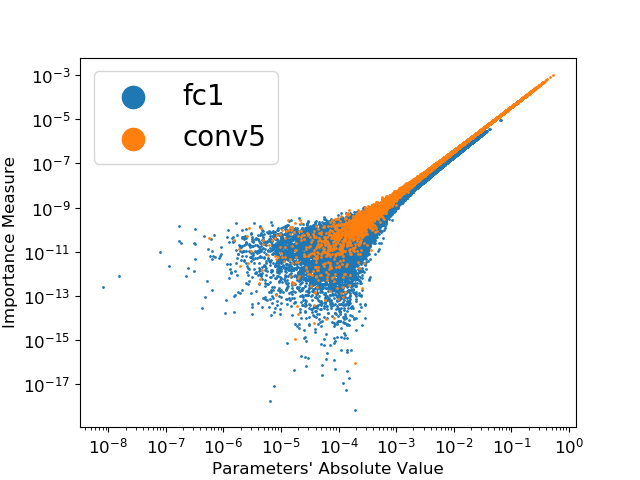}
        \caption{25\% prune to 12.5\%}
        \label{fig:25to12.5}
    \end{subfigure}
    \caption{Correlation between parameters' absolute values and importances.}
    \label{fig:correlation}
\end{figure*}

We further explore the correlation between a parameter's magnitude and its importance measured by NAP. From figure \ref{fig:correlation}, we can observe that: 1) A parameter's absolute value indeed has a strong correlation with its importance computed by NAP. This explains why magnitude-based methods also achieve fairly good performance. Despite this strong correlation, directly compare the magnitudes from all layers will a give different pruning result than our method: setting the pruning threshold as the median of parameters' magnitudes in Figure \ref{fig:100to50} will prune 14\% of conv5, while the median of our importance will prune 7\%. Therefore, directly compare magnitudes across all layers and prune the smallest ones will lead to a sub-optimal architecture, and limit the performance, and thus magnitude-based methods prune layers individually. 2) The FC layer initially has smaller importance(larger redundancies) than convolutional layer measured by NAP, and thus are pruned more severely.
As more and more parameters are pruned away(from Figure \ref{fig:100to50} to Figure \ref{fig:25to12.5}), the importance of FC layer becomes closer to that of convolutional layer, and both layers will be pruned with similar ratios. This shows that our method can dynamically adjust the compression ratio of each layer, as the pruning going on.

\section{Does NAP sensitive to hyper-parameters?}
\label{app:hyper}
As already demonstrated in the ablation study, our method is robust to the choice of hyper-parameters. NAP has two hyper-parameters: the pruning ratio $p$ at each pruning step, and the number of steps for fine-tuning (updating stats) $T$\footnote{Since we update stats and fine-tune the network simultaneously, we use those to terms interchangeably}. In the ablation study, we show that within a fixed pruning time-budget (the total number of steps run on the training dataset before achieving $5.4$ times acceleration), different choices of $p$ and $T$ gives similar final performance. We believe the fixed pruning time-budget is a practical constraint, as one wants to get a small network as soon as possible. Here, we first show more results with fixed time-budget, and then show the effect of other hyper-parameter choices without this constraint.

\begin{figure*}[htb]
    \centering
        \centering
        \includegraphics[height=7in]{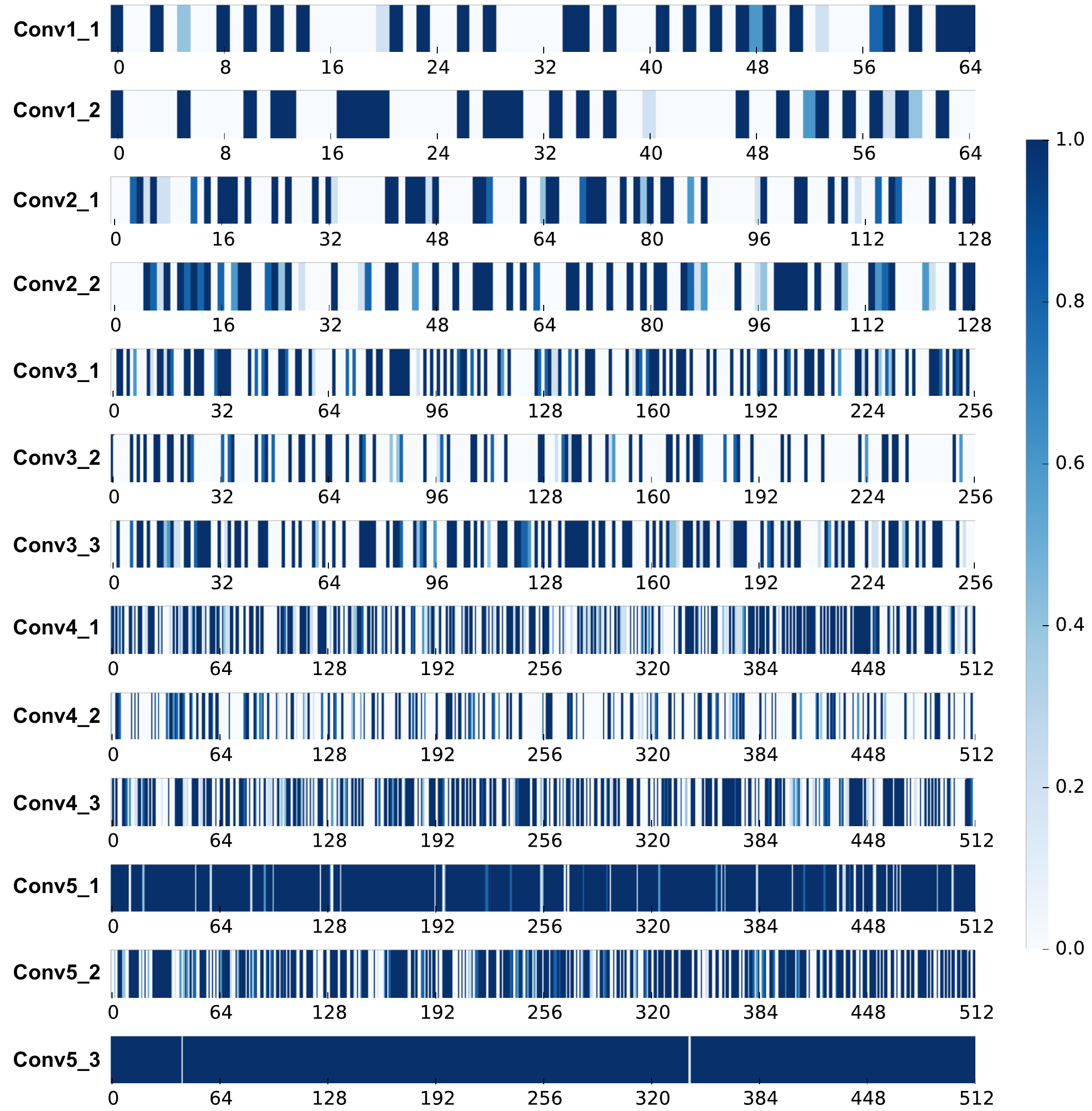}
        \caption{Remained channel index averaged over 5 experiments with different hyper-parameters. Dark blue means a channel is remained in all 5 experiments, while white indicates a channel is pruned in all 5 experiments. This heatmap shows that though using different hyper-parameters, our method ends up with consistent model architectures.}
        \label{fig:channel_id}
\end{figure*}

In addition to similar performances, we also notice that the final model architectures given by different hyper-parameters are very similar. Figure \ref{fig:channel_id} shows the remained channel index in a pruned VGG16, averaged over experiments using different hyper-parameters. Same as before, we conduct 5 different channel-wise pruning experiments, with 1) $p=10\%, T=\text{2 epoch}$, 2) $p=5\%, T=\text{1 epoch}$, 3) $p=2\%, T=\text{0.4 epoch}$, 4) $p=1\%, T=\text{0.2 epoch}$, 5) $p=0.5\%, T=\text{0.1 epoch}$. These give us 5 pruned VGG16 network, with similar final performance (as shown in the ablation study). We then average over these 5 architectures on whether a channel is pruned or not. For instance, if a specific channel is pruned in all 5 models, then its corresponding value is 0, if it's pruned in 4 models, then its value is 0.2, and if it's not pruned in all model, then its value is 1. Based on these values, we plot the heat-map in Figure \ref{fig:channel_id}.

\begin{figure}[htb]
    \centering
        \centering
        \includegraphics[height=2in]{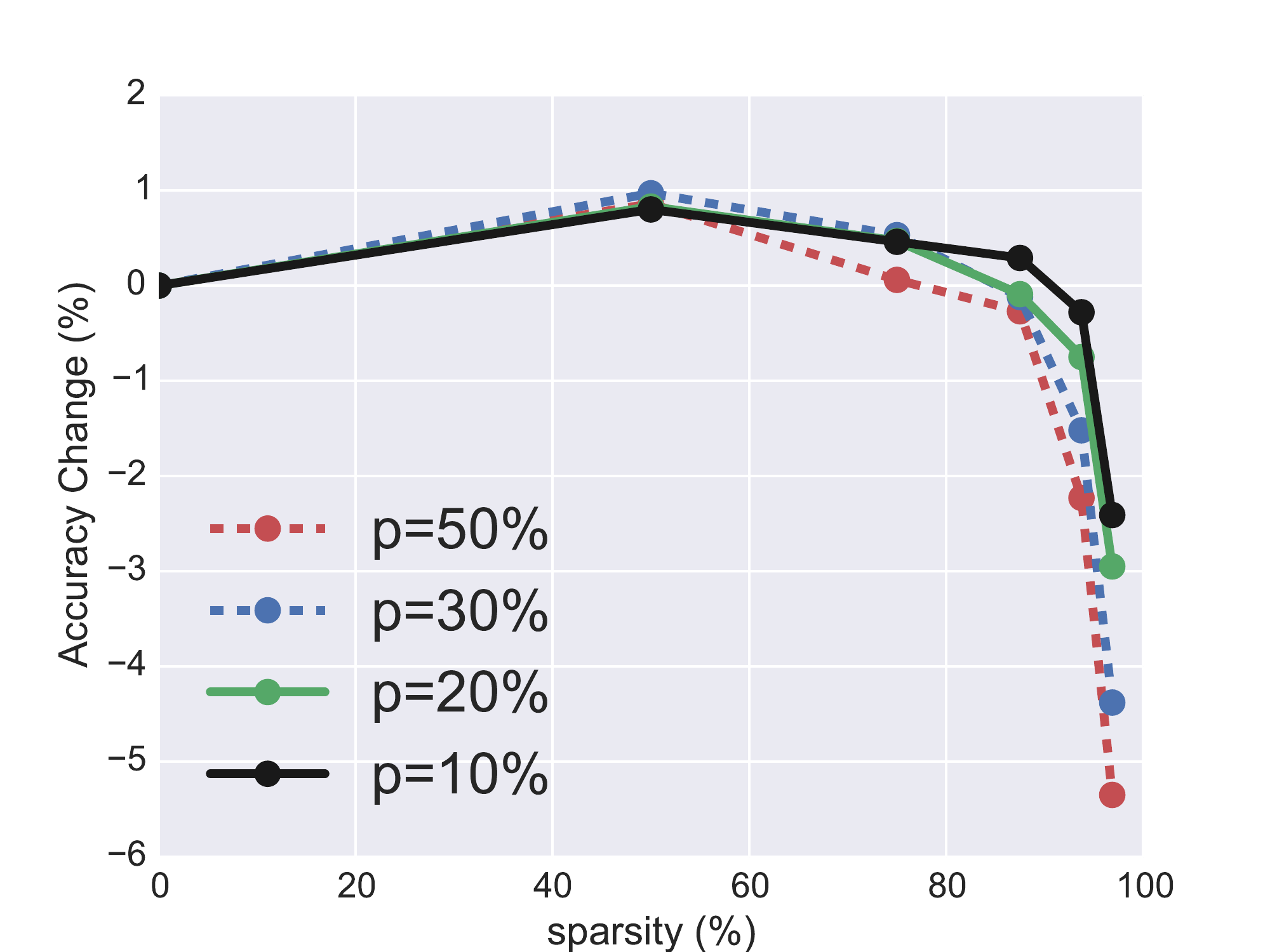}
        \caption{Different value of pruning fraction $p$. Higher sparsity means fewer parameters.}
        \label{fig:different_p}
\end{figure}
\begin{figure}[htb]
        \centering
        \includegraphics[height=2in]{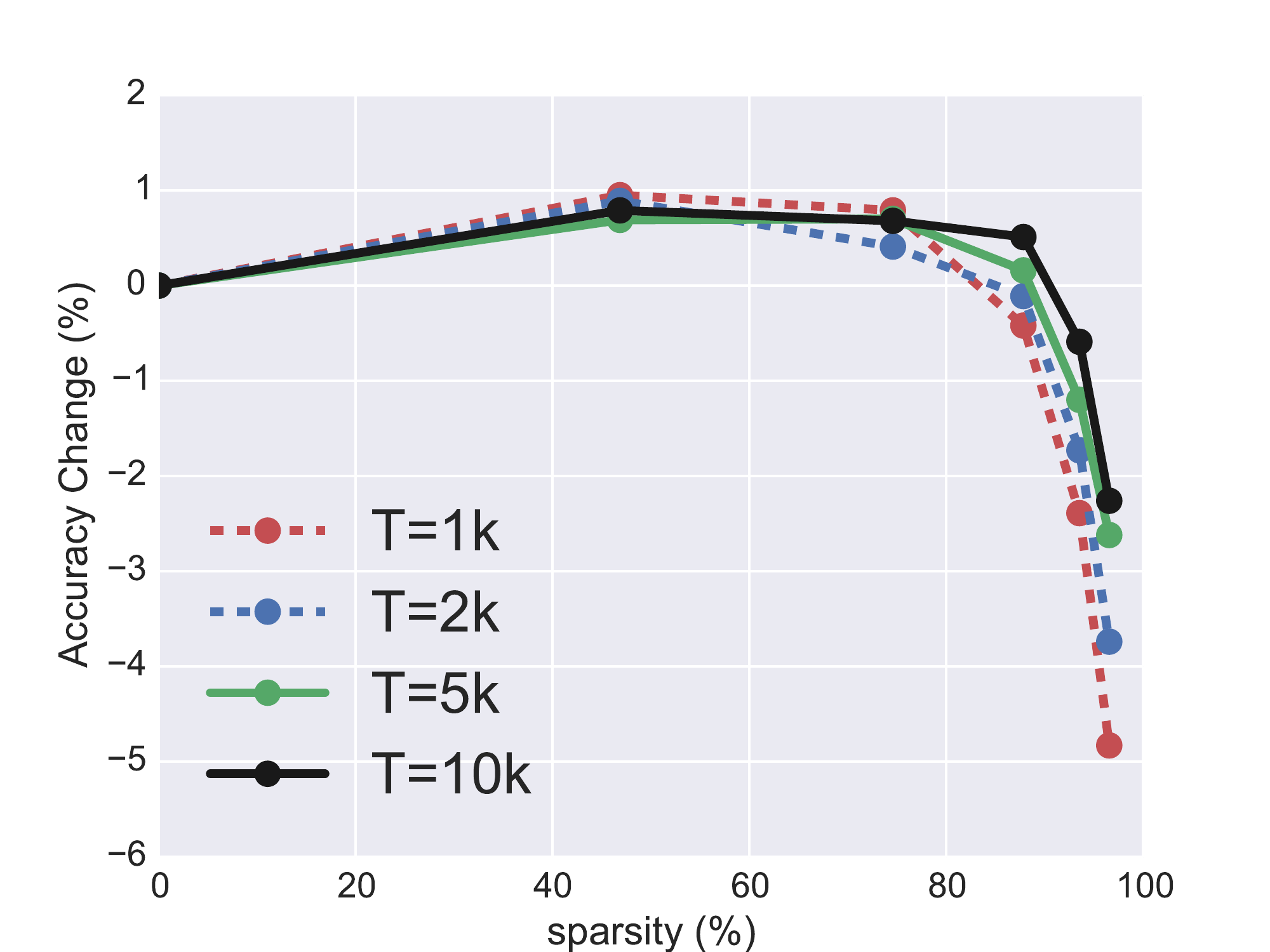}
        \caption{Different value of fine-tuning step $T$. Higher sparsity means fewer parameters.}
        \label{fig:different_s}
\end{figure}

From Figure \ref{fig:channel_id} we can see that the pruned models have very consistent architectures. Channels generally have dark blue or white color, meaning they are either maintained in all 5 models or pruned in all 5 models. This presented heat-map is very different from a random pruning pattern, in which channels has 0.5 probability of either pruned or maintained, and thus the associated heatmap will have a intermediate blue color for all channels. This experimental result further prove that our method is robust to the choice of hyper-parameter. Since different hyper-parameters end up with similar architecture, the final performance are similar will be a reasonable result. Furthermore, all these 5 pruned model originated with the same pre-trained model. We believe the consistent architectures also indicates that there may exist a one-one correspondance between a set of pre-trained weights and a effective pruned architecture. We hope this observation will lead to more discussions in future work.

Next, we show experimental results with other hyper-parameter choices, regardless of the fixed pruning time-budget. These include effects of different $p$ and effects of different $T$. Intuitively, smaller $p$ should give better results, as we will have a more accurate approximation of the Hessian matrix. Here, we apply our method with fixed $T$ and different $p$ in the {\it fine-grained setting}, using CifarNet. %
Figure \ref{fig:different_p} shows the relation of final accuracy under different sparsity levels. In particular, we apply different $p$ followed by 20k steps of fine-tuning. We can see that small $p$  has better results, especially when the sparsity is larger than 90\%. On the other hand, there are negligible differences in a broad range of sparsity, i.e. sparsity is less than 90\%. Therefore, we believe the choice of $p$ will depends on one's time-budget for compression; smaller $p$ will give better result but consume longer time, while larger $p$ will still give relatively good results using much shorter time.

We also have similar conclusions for different values of $T$. Since we use the Fisher matrix as a proxy for the  Hessian, we assume the model is converged before each pruning. Thus, longer fine-tuning steps will help estimate a more accurate Hessian matrix. As shown in Figure \ref{fig:different_s}, larger $T$ yields better results. Moreover, the value of $T$ has almost no effect when sparsity is small. Similar to $p$, we believe the choice of $T$ also depends on the time-budget for compression, and it is not hard to tune.

\begin{table*}[htb]
  \centering
  \begin{tabular}{l|l|llll}
    \hline
    \hline
    Architecture & Layer & Parameters & LWC\cite{han2015learning} & DNS\cite{guo2016dynamic}& \textbf{NAP} (Ours)\\
    \hline
    \multirow{4}{*}{LeNet-300-100}
    & fc1 & 235K & 8\% & 1.8\% & 0.73\%\\
    & fc2 & 30K & 9\% & 1.8\% & 4.74\%\\
    & fc3 & 1K & 26\% & 5.5 \% & 39.4\% \\
    \cline{2-6}
    & total & 267K & 8\% & 1.8\% & \textbf{1.3\%}\\
    \hline
    \multirow{ 5}{*}{LeNet-5}
    & conv1 & 0.5K & 66\% & 14.2\% & 41.2\%\\
    & conv2 & 25K & 12\% & 3.1\% & 3.1\%\\
    \cline{2-6}
    & fc1 & 400K & 8\% & 0.7\% & 0.2\%\\
    & fc2 & 5K & 19\% & 4.3\% & 8.9\% \\
    \cline{2-6}
    & total & 431K & 8\% & 0.9\% & \textbf{0.5\%}\\
    \hline
    \multirow{ 9}{*}{AlexNet}
    & conv1 & 35K & 84\% & 53.8\% & 67.2\%\\
    & conv2 & 307K & 38\% & 40.6\% & 37.8\%\\
    & conv3 & 885K & 35\% & 29.0\% & 27.7\%\\
    & conv4 & 663K & 37\% & 32.3\% & 33.2\%\\
    & conv5 & 442K & 37\% & 32.5\% & 38.6\%\\
    \cline{2-6}
    & fc1 & 38M & 9\% & 3.7\% & 1.5\%\\
    & fc2 & 17M & 9\% & 6.6\% & 3.2\%\\
    & fc3 & 4M & 25\% & 4.6\% & 13.7\%\\
    \cline{2-6}
    & total & 61M & 11\% & 5.7\% & \textbf{4\%}\\
    \hline
    \multirow{ 17}{*}{VGG16}
    & conv1\_1 & 2K & 58\% & - & 92.1\%\\
    & conv1\_2 & 37K & 22\% & - & 66.4\%\\
    \cline{2-6}
    & conv2\_1 & 74K & 34\% & - & 55.7\%\\
    & conv2\_2 & 148K & 36\% & - & 46.9\%\\
    \cline{2-6}
    & conv3\_1 & 295K & 53\% & - & 38.2\%\\
    & conv3\_2 & 590K & 24\% & - & 30.8\%\\
    & conv3\_3 & 590K & 42\% & - & 30.5\%\\
    \cline{2-6}
    & conv4\_1 & 1M & 32\% & - & 20.6\%\\
    & conv4\_2 & 2M & 27\% & - & 12.9\%\\
    & conv4\_3 & 2M & 34\% & - & 12.8\%\\
    \cline{2-6}
    & conv5\_1 & 2M & 35\% & - & 13.1\%\\
    & conv5\_2 & 2M & 29\% & - & 13.9\%\\
    & conv5\_3 & 2M & 36\% & - & 13.3\%\\
    \cline{2-6}
    & fc1 & 103M & 4\% & - & 0.3\%\\
    & fc2 & 17M & 4\% & - & 1.9\%\\
    & fc3 & 4M & 23\% & - & 9.2\%\\
    \cline{2-6}
    & total & 138M & 7.5\% & - & \textbf{4\%}\\
    \hline
    \hline
  \end{tabular}
  \caption{{\it Fine-grained} compression ratios for different layers.}
  \label{table:per_ratio}
\end{table*}

\section{Does NAP find reasonable per-layer compression ratios?}
\label{sec:exp_ratios}
We  compare the {\it fine-grained} compression ratios, layer by layer, between our method and previous methods, as shown in Table \ref{table:per_ratio}. The compression ratios of LWC \cite{han2015learning} and DNS \cite{guo2016dynamic} are manually tuned to achieve good results. In contrast, our method automatically determines the compression ratio for each layer during the pruning process. We notice that our compression ratios are not the same as previous manually-tuned ratios, but share some similarities; layers with smaller compression ratios in our method also have smaller ratios in other methods. Also, fully-connected layers are generally be pruned much severely than convolutional layers, which is in accord with the observation that fully-connected layers usually have more redundancies. These suggest that our method can find reasonable per-layer compression ratios according to the sensitivities of each layer.

\section{Implementation Details}
Here, we provide our more detailed implementation details for future reproduction. Our implementation of NAP is based on Tensorflow. On imagenet, we first pretrain models following hyper-parameter settings in Caffe model zoo (AlexNet and VGG16) or the original paper (ResNet50). Our pretrained model match the performance in Caffe model zoo (or original paper for ResNet50). Through all training pipeline, the training augmentation is first resize the shortest side to a random size, and then randomly crop a 224x224 image with random horizontally flip (except for AlexNet we resize to a fixed size of 256 and randomly crop a 227x227 image). For testing, we first resize the shortest side to 256, and then center crop a 224x224 image. No other augmentation is used. For all experiments, we use SGD with momentum of 0.9.

NAP has an operation to update second-order stats of input activations and output gradients. Our implementation of this is based on the open sourced K-FAC repository,\footnote{https://github.com/tensorflow/kfac} with some modifications. Recall that we can fine-tune the network while we update those stats. However, updating the stats typically consume many memories and encounter Out-Of-Memory issue when conducting on GPUs. Therefore, for large network like VGG, we update the stats on CPU. Moreover, the update stats operation is conducted asynchronously with the forward-backward pass on GPU, since this will hide the overhead and save the time. When doing inverse, we also add a damping term to increase the numerical stability.

For channel-wise pruning, we remove $1\%$ of the remaining channels from a pre-trained model, and fine-tune 0.2 epoch between two subsequent pruning. In total, we run less than 8 epochs before we achieve a 5.4x accelerated VGG16. This time budget is similar or shorter than previous methods. For the last stage of fine-tuning VGG, we use 1e-3 as the initial learning rate, decay to 1e-4 at 20 epochs and stop at 30 epochs. The weight decay term is set to 0. For ResNet, we use 1e-3 as the initial learning rate, decay by 10 at 10 epochs and 20 epochs, and stop at 30 epochs. In our experiments, we find that we can achieve better performance if we train with longer time. However, the goal of the experiments is to show the effectiveness of our method, rather than the final fine-tuning schedule. Therefore, we keep our fine-tune setting similar as previous work.

For the fine-grained pruning, we prune more aggressively at each stage since a network is less sensitive to fine-grained pruning. More specifically, we prune AlexNet and VGG16 iteratively to $50\%$, $25\%$, $12.5\%$, $6.25\%$, $5\%$, $4\%$. Such compression ratios are chosen because we simply choose to halve the network size at the very beginning, and then prune less aggressively with $80\%$ compression ratio in the last two pruning steps. After each pruning, we retrain the network with the same learning schedule as in the pre-training, with slightly lower weight decay (2e-4). We empirically find that such schedule can converge faster than fine-tune with a small learning rate (e.g. 1e-4, which can converge to a similar performance but with much longer time: ~200 epochs as shown in \cite{han2015learning}). For ResNet50, we follow similar setting in \cite{he2018amc}. We iteratively prune ResNet50 to $50\%$, $30\%$, $25\%$, $20\%$, $15\%$. The fine-tune learning schedule is simply use 1e-4 and train until the model converges.

\end{document}